\documentclass[graybox]{svmult}

\usepackage{algorithm,algorithmic}
\usepackage{amsmath}  
\usepackage{tikz}
\usepackage{mathptmx}       
\usepackage{helvet}         
\usepackage{courier}        
\usepackage{type1cm}        
\usepackage{makeidx}         
\usepackage{graphicx}        
\usepackage{multicol}        
\usepackage[bottom]{footmisc}
\usepackage[colorinlistoftodos]{todonotes}

\newcommand{\ttA}{\textbf{\texttt{a}}}
\newcommand{\ttB}{\textbf{\texttt{b}}}
\newcommand{\ttC}{\textbf{\texttt{c}}}
\newcommand{\ttD}{\textbf{\texttt{d}}}
\newcommand{\ttE}{\textbf{\texttt{e}}}

\usetikzlibrary{matrix}
\newcommand{\ing}[1]{\includegraphics[width=\linewidth]{#1}}
\tikzset{
  img/.style={
    text width=0.325\textwidth,
    inner sep=0pt,     
    outer sep=0pt,     
    rectangle,
    align=center,
    anchor=west,
    draw=none} 
}

\makeindex             

\begin{document}

\title*{Distributed Camouflage for Swarm Robotics and Smart Materials}
\author{Yang Li, John Klingner, and Nikolaus Correll}
\institute{Yang Li, John Klingner, and Nikolaus Correll\at Department of Computer Science, University of Colorado, Boulder, USA \\
\email{yang.li-4@colorado.edu, john.klingner@colorado.edu, nikolaus.correll@colorado.edu}
\and Nikolaus Correll \at URL: http://correll.cs.colorado.edu/
}
%
%
\maketitle

\abstract{
We present a distributed algorithm for a swarm of active particles to camouflage in an environment. Each particle is equipped with sensing, computation and communication, allowing the system to take color and gradient information from the environment and self-organize into an appropriate pattern. Current artificial camouflage systems are either limited to static patterns, which are adapted for specific environments, or rely on back-projection, which depend on the viewer's point of view. Inspired by the camouflage abilities of the cuttlefish, we propose a distributed estimation and pattern formation algorithm that allows to quickly adapt to different environments. We present convergence results both in simulation as well as on a swarm of miniature robots ``Droplets'' for a variety of patterns.   
}

\section{Introduction}
We wish to design artificial camouflage systems that can quickly adapt to a large variety of environments. Inspired by the capabilities of cephalopods, which tightly integrate sensing, actuation (color change), neural computation and communication, we are interested in a distributed artificial approach that mimics this tight integration \cite{yu2014adaptive,mcevoy2015materials}. While animals employ camouflage mostly for escaping predators, camouflage in an engineering context is typically motivated by clandestine military operations. More broadly, everything from small robots to buildings could use these techniques to more seamlessly be a part of their environment. Nature employs a large variety of techniques to achieve these goals. For example, moths mimic patterns that they would expect in their environments, sea animals use mottle patterns to soften their contours, and other animals decorate their body with artifacts from the environment \cite{stevens2009animal}. Two animals with notable camouflage abilities are the cuttlefish and octopus, who can dramatically alter the coloration and patterning of their skin and switch between different environments in a matter of seconds \cite{hanlon1988adaptive}. These creature's camouflage behavior is not only driven by the animals' visual system (which is color-blind \cite{messenger1977evidence}) or brain \cite{messenger2001cephalopod}, but has also been shown to rely on local sensing and control \cite{ramirez2015eye}. 

There have been multiple attempts to achieve active camouflage using a combination of cameras and projection \cite{inami2003optical,lin2009framework}. Although such systems provide ``perfect'' camouflage, they are highly dependent on the observer's viewpoint. Mimicking the background exactly is rarely employed in the animal kingdom, where a few simple families of patterns --- mottled, striped, or simply uniform \cite{hanlon2007cephalopod} --- dominate. Creating such patterns requires only local coordination \cite{meinhardt1982models}, suggesting a combination of high-level selection of appropriate motor programs \cite{messenger2001cephalopod} and self-organization \cite{meinhardt1982models}. Here, we are not concerned with perfectly matching the background, but rather aim to replicate the pattern matching ability of natural systems, which are able to fool sophisticated predators.

Distributing the sensing and actuation for camouflage generation makes an implementation scalable for a variety of factors, such as resolution of the camouflage pattern, the size of the area being camouflaged, and robustness against the failure of individual units. Further, a distributed camouflage system could respond to local changes in the environment, in particular when deployed on non-trivial 3D surfaces. 

In this paper, we present a fully distributed approach, which we implement on a swarm of Droplets \cite{farrow2014miniature}, each equipped with the ability to sense and emit color as well as communicate with its local neighbors. Although there exist multiple attempts to design artificial chromatophores, most work focuses on component technology, i.e. the ability to color change in a soft substrate \cite{rossiter2012biomimetic,morin2012camouflage}, but very few works articulate the systems challenges that require not only local color changes, but also local sensing and computation \cite{yu2014adaptive}, or investigate the ability to co-locate simple signal processing with the sensors themselves \cite{fekete2009distributed}.  



Our algorithm can be broken into three phases, each described in detail below. First, we estimate a color and gradient histogram with a consensus algorithm among the particles. This information is then used to determine the parameters of a pattern formation algorithm. Finally, the pattern is formed using a reaction-diffusion process. 
The ``background'' in to which the swarm is trying to camouflage is projected on to the particles from above, requiring them to have color sensors. To simplify the color-identification process, the paper focuses on two-tone patterns. Finally, we arrange the particles in a grid pattern, which allows us to implement a discrete convolution operation and simplifies debugging the pattern at the low resolution that swarms in the order of tens of particles can afford.

\section{Distributed Camouflage Algorithm}
\label{sec:camouflage}

\begin{figure}[!htb]
\centering
\includegraphics[width=\columnwidth]{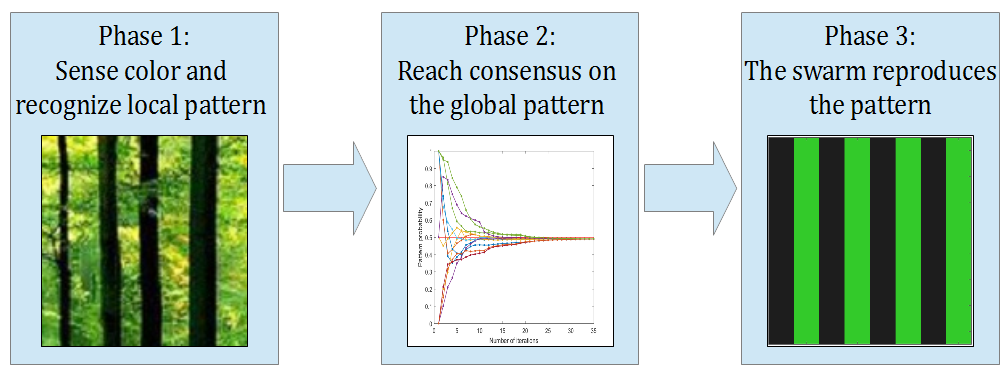}
\caption{Pipeline of the Distributed Camouflage Algorithm }
\label{fig:pipeline}
\end{figure}

In this section, we describe the distributed camouflage algorithm. Fig.~\ref{fig:pipeline} illustrates the steps of the algorithm in broad strokes. First, each robot measures the color projected on it. Then, it exchanges the measured color with neighboring robots. Once received, neighbors' color information is used to compute an estimated probability for the various pattern types based on local information (Section~\ref{sec:descriptor}). Next, the swarm communicates their local pattern probabilities, using a weighted-average consensus algorithm to compute the most likely global pattern (Section~\ref{sec:consensus}). Once consensus has been achieved, the swarm reproduces the pattern collaboratively with a reaction-diffusion process (Section~\ref{sec:generator}).

\subsection{Pattern Descriptor}
\label{sec:descriptor}

\begin{figure}[!htb]
\sidecaption[t]
\includegraphics[width=0.5\columnwidth]{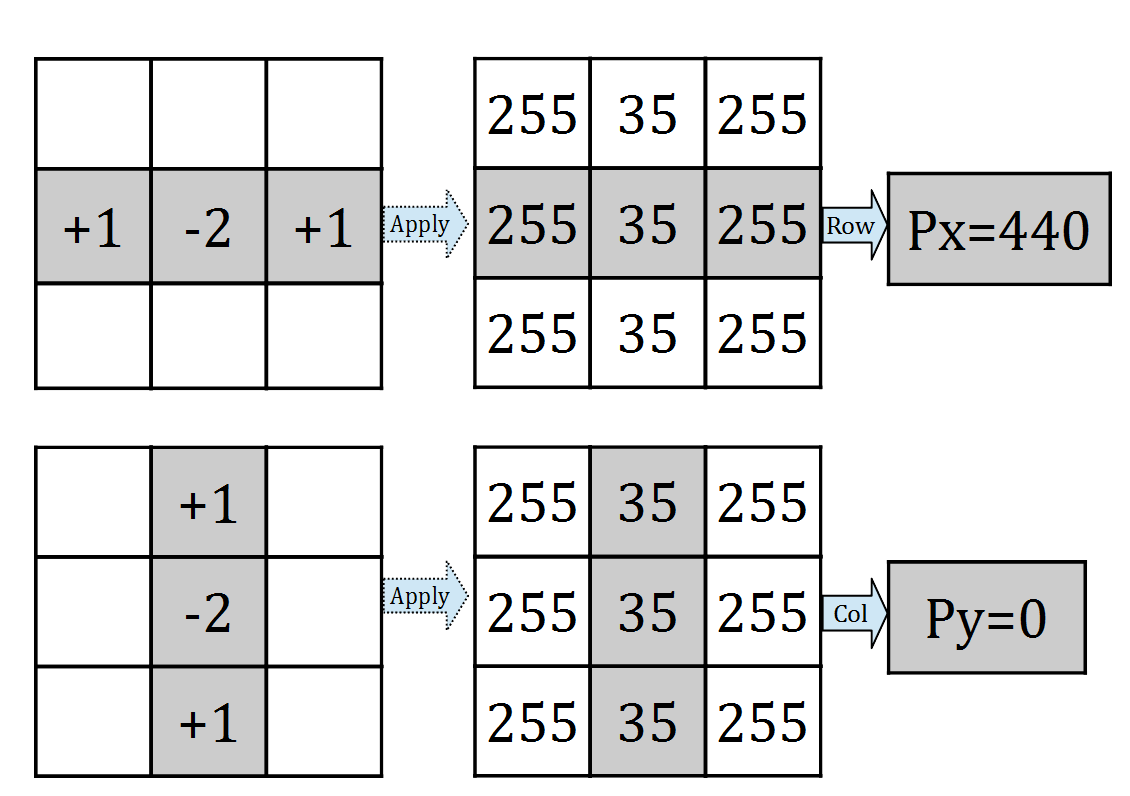}
\caption{Illustration of applying the two second order derivative masks.}
\label{fig:filter}
\end{figure}

Once each robot has measured the local environment's color, and communicated that information, they apply a filter mask (see Fig.~\ref{fig:filter}) to compute a discrete approximation of the second-order color derivative in both the horizontal and vertical directions. This is quite similar in concept to kernels used in edge detection and other computer-vision tasks~\cite{dalal2005histograms}. Indeed, if the grid of robots is viewed as an image with each robot a pixel, these two pattern descriptors are simply the value the pixel would have after each of the two convolutions. These second-order derivatives are the \emph{Pattern Descriptors} -- denoted $P_x$ and $P_y$ for the horizontal and vertical directions respectively -- and are used to calculate the most probable local pattern.

To be specific, with $M$ denoting my local color and $T$, $R$, $B$, and $L$ denoting the color of my top, right, bottom, and left neighbors, $P_x$ and $P_y$ are given by:
\begin{align*}
P_x &= L + R - 2M \\
P_y &= T + B - 2M
\end{align*}
A pattern-probability array $p = [p_h, p_v, p_m]$ is used to record each robot's pattern, where $p_h$ represents the probability of a  pattern-type is selected and given a probability of $1$ based on our local \emph{Pattern Descriptors}, and the other probabilities are all 0. This is shown in the equation below, where $T$ is some threshold value and $\left|val\right|$ is used to indicate $\texttt{abs}\left(val\right)$.
\begin{align}
\label{eq:pattern_descriptor}
p = [p_h, p_v, p_m] = 
\begin{cases} 
[1, 0, 0] & \text{if } \left|P_y\right|-\left|P_x\right|>T \\ 
[0, 1, 0] & \text{if } \left|P_x\right|-\left|P_y\right|>T \\ 
[0, 0, 1] & \text{otherwise} \end{cases}
\end{align}
Note that a grid representation has only been chosen for the simplicity of performing (and explaining) the mathematical operations, but one could equally well perform the described convolutions using continuous representations and local range and bearing information. 

\subsection{Distributed Average Consensus Scheme}
\label{sec:consensus}

Once each robot has computed the most likely local pattern (ie, computed $p = [p_h, p_v, p_m]$), they need to achieve consensus on the global pattern. We use the distributed average consensus scheme~\cite{xiao2005scheme} for this purpose. In each step of this scheme, the robot updates its local $p$ to be a weighted average of its own and its neighbors'. This step is repeated many times, allowing information to diffuse through the swarm. Since the weighted average just uses local information, each step takes the same amount of time regardless of the number of robots in the swarm. The number of steps needed was determined experimentally. 


The weighted-average calculation uses Metropolis weights, defined as:
\begin{equation}
\label{eq: weights}
W_{i,j} =
\left\{
    \begin{array}{ll}
    \frac{1}{1+max\{ d_i, d_j \}}& \text{if } (i, j) \in E,\\
    1-\sum_{(i,k) \in E} {W_{i,k}}& \text{if } i = j,\\
    0& \text{ otherwise.}
    \end{array}
\right.
\end{equation}
The Metropolis weights are well-suited for distributed algorithms, since weight-calculation requires only local knowledge. Further, it is proven in~\cite{xiao2005scheme} that Metropolis weights guarantee convergence of the average consensus provided that the infinitely occurring communication graphs are jointly connected. Once the robots have converged, the largest value in $p$ represents the most likely global pattern. For example, $p_h > p_v$ and $p_h > p_m$ indicates that the most likely global pattern is horizontal stripes.


\subsection{Pattern Generator}
\label{sec:generator}

In this section, we describe the distributed pattern formation algorithm to generate a proper pattern to match the environment.

\begin{figure}[!htb]
\centering
\includegraphics[width=0.8\columnwidth]{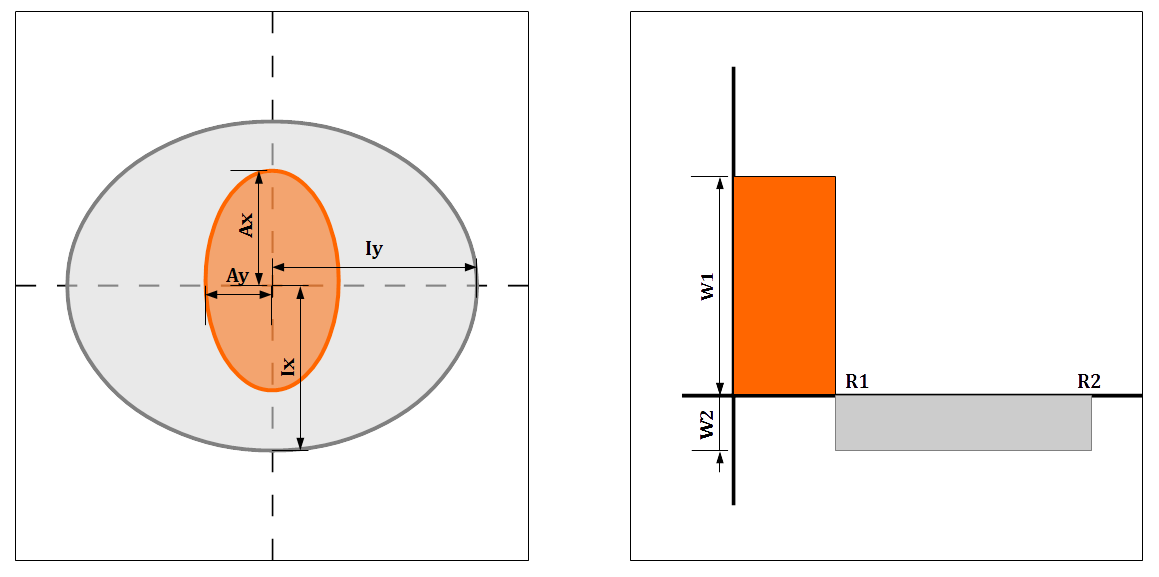}
\caption{Illustration of local activator-inhibitor model: on the left, the activation region (orange) is defined by $A_x$ and $A_y$ while the inhibition region (gray) is defined by $I_x$ and $I_y$; on the right, $W_1$ and $W_2$ are the two field values. $R1$ is related to $A_x$ and $A_y$ and $R2$ is related to $I_x$ and $I_y$, }
\label{fig:morphogen}
\end{figure}

Now that a global pattern has been selected, the robots next need to generate an appropriate camouflage pattern. We use the pattern-formation algorithm presented by Young~\cite{young1984local}: a local activator-inhibitor model. In this model, each cell (robot) is either `on' or `off', and can generate two kinds of morphogens: activator morphogen and inhibitor morphogen. Together, these form a ``morphogenetic field''. Note that the activator should be inside of the inhibitor (see left of Figure~\ref{fig:morphogen}). The cells (robots) in the activator morphogen contribute to stimulate change for nearby `on' cells, and cells in the inhibitor morphogen contribute to stimulate change for nearby `off' cells.

\begin{figure}[!htb]
\centering
\includegraphics[width=0.95\columnwidth]{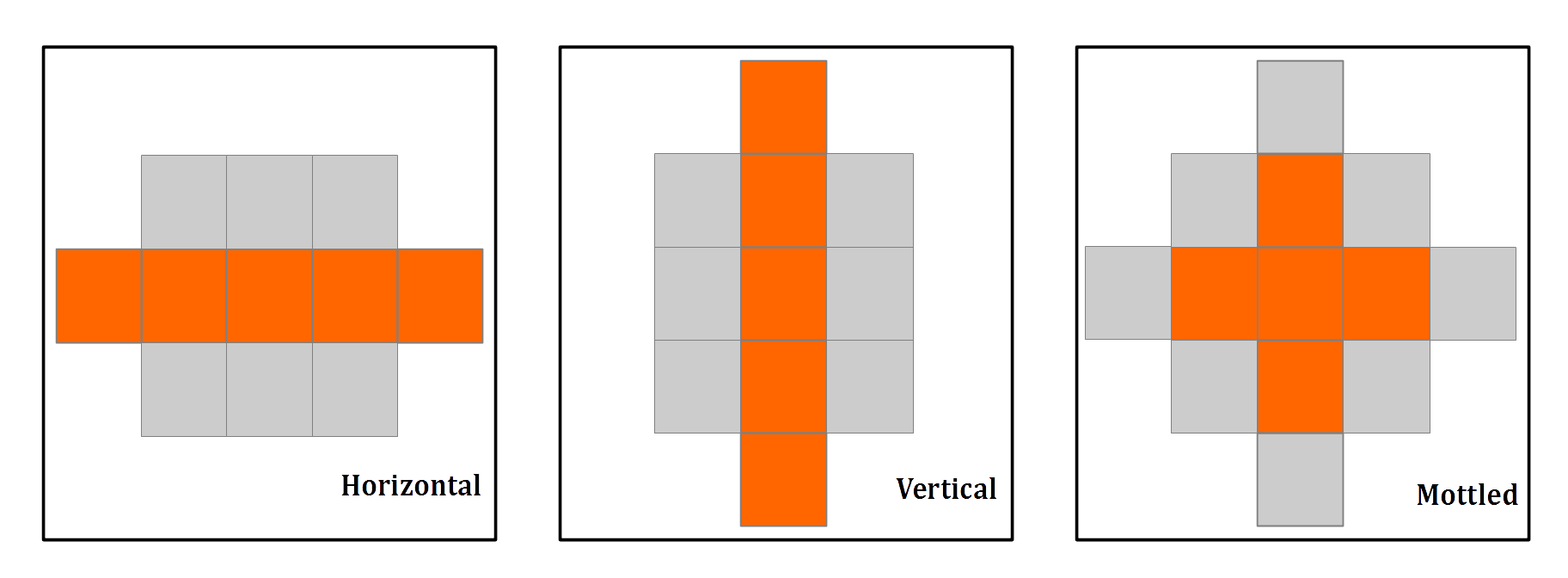}
\caption{Activator (orange) and Inhibitor (gray) Regions for each of the three patterns.}
\label{fig:droplet_morphogen}
\end{figure}

During each step of this algorithm, each cell changes its `off'/`on' status based on the combined effect of all nearby morphogenetic fields. More specifically, a `strength' is calculated with each `on' robot contributing a positive value ($W_1$) if in the activator region or contributing a negative value ($W_2$) if outside the activator region and inside the inhibitor region. The robot then changes its state to `on' if the strength is greater than 0, and to `off' otherwise. This step is repeated until the states converge to a stable pattern. In~\cite{young1984local} the author observes that convergence typically takes around five steps. This was consistent with our observations.

In this framework, the different types of patterns are represented with differently shaped activator and inhibitor regions. The regions for each pattern are shown in Figure~\ref{fig:droplet_morphogen}. Note that the region sizes mean that each robot only requires information from robots within two hops of it.



\section{Simulated Results}
\label{sec:simulation}
We implemented the algorithm introduced above on a centralized system for testing. By presenting some simulated results here, we hope to demonstrate the algorithm's functionality and add clarity to the explanation above. We run these tests with three images, one for each of the pattern types. Each image is $128 \times 128$ pixels, and gray scale. We simulate $64$ ($8 \times 8$) robots.

\begin{figure}
\begin{tikzpicture}[trim left = -4.75cm]
  \matrix[matrix of nodes,nodes=img, column sep=0cm]{
    |[label={[label distance=-1.6cm]180:{\Large \ttA}}]| \ing{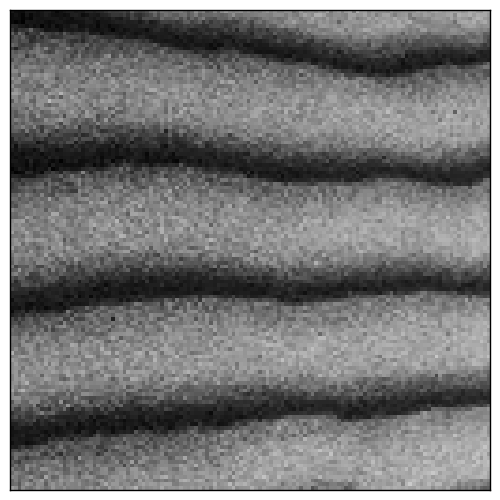} & \ing{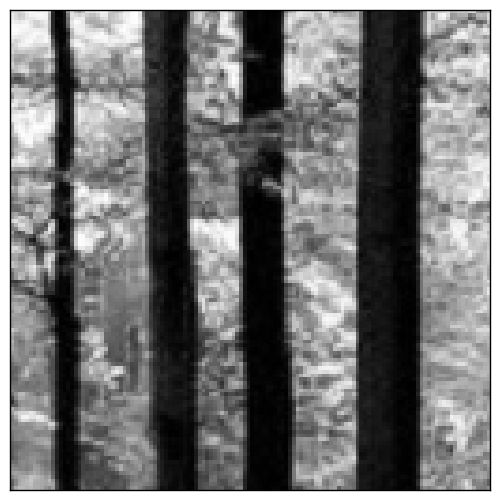} & \ing{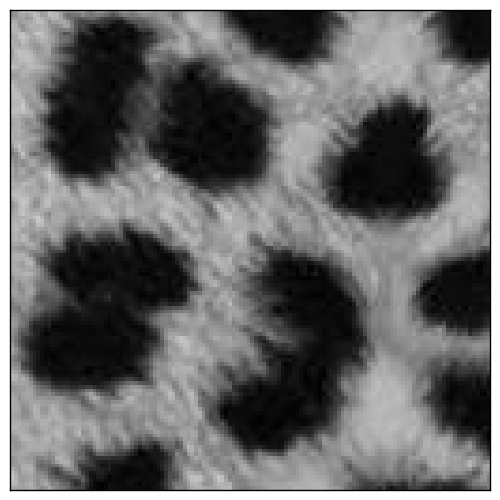} \\
    |[label={[label distance=-1.6cm]180:{\Large \ttB}}]| \ing{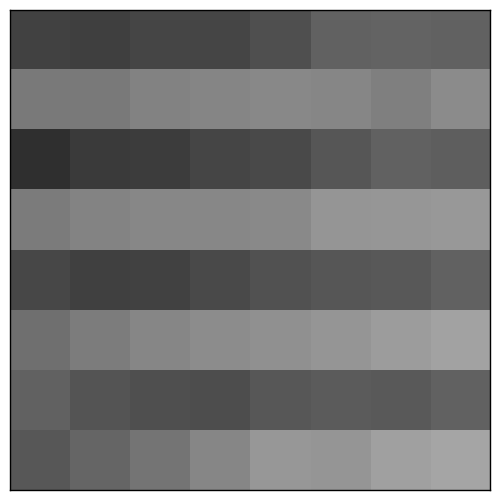} & \ing{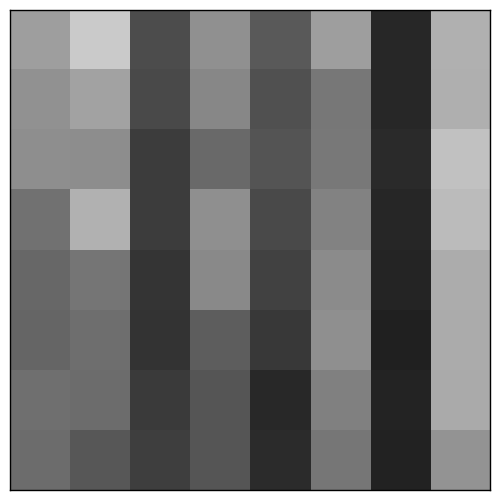} & \ing{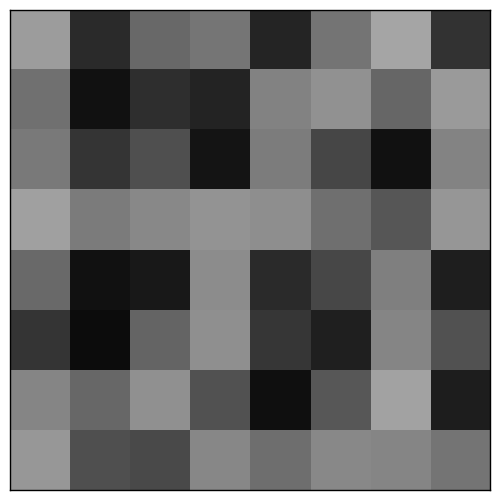} \\
    |[label={[label distance=-1.6cm]180:{\Large \ttC}}]| \ing{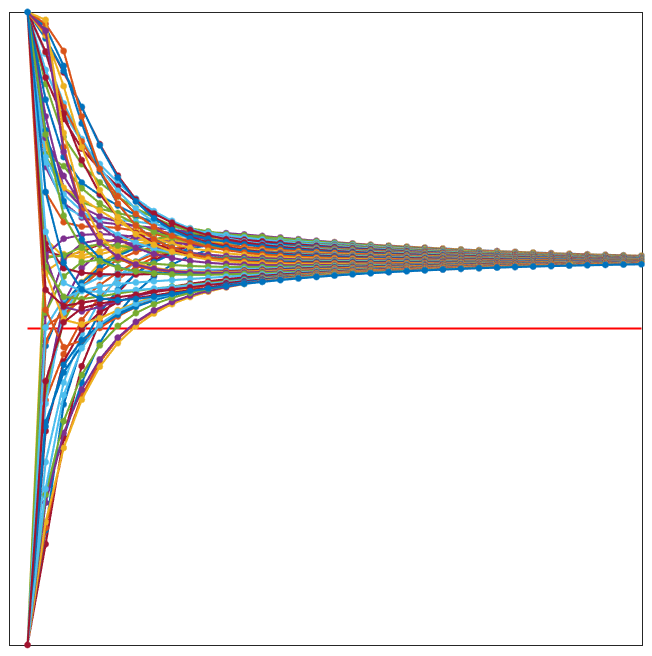} & \ing{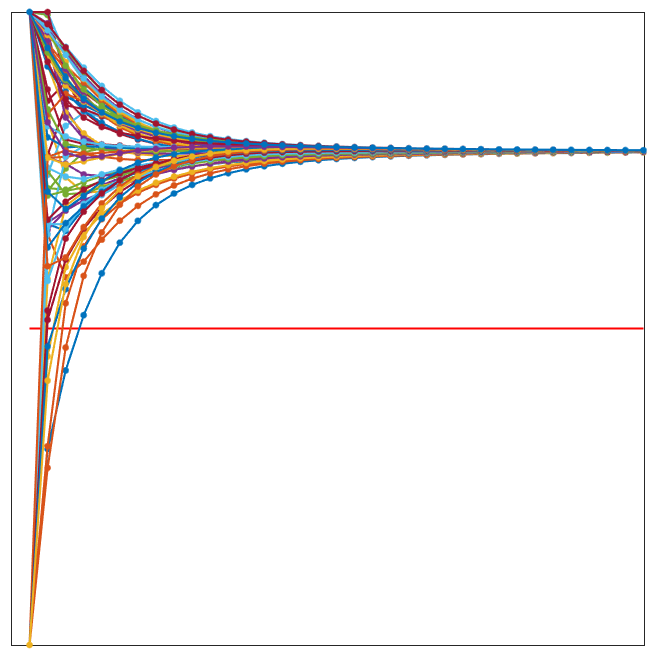} & \ing{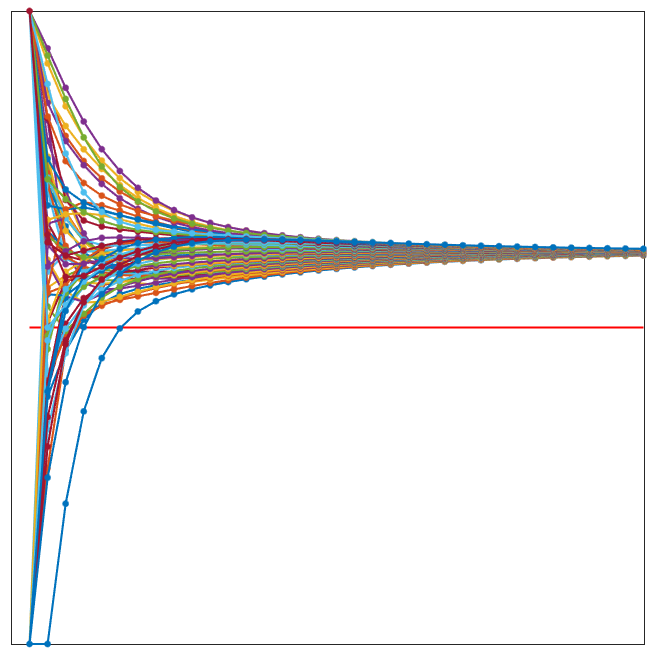} \\
    |[label={[label distance=-1.6cm]180:{\Large \ttD}}]| \ing{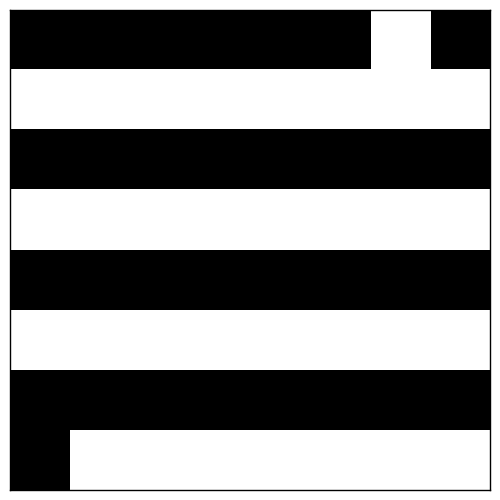} & \ing{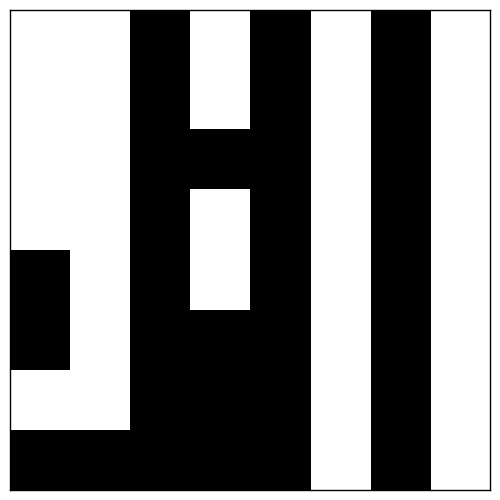} & \ing{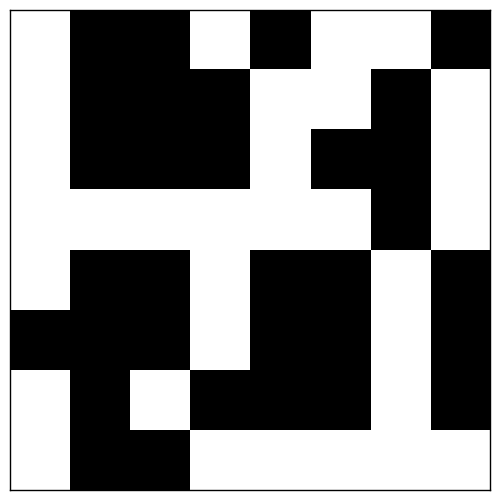} \\
    |[label={[label distance=-1.6cm]180:{\Large \ttE}}]| \ing{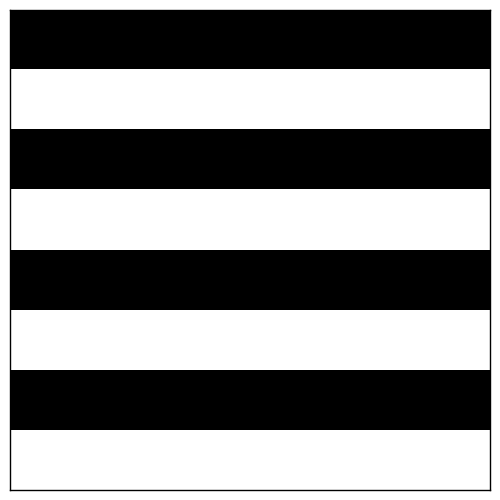} & \ing{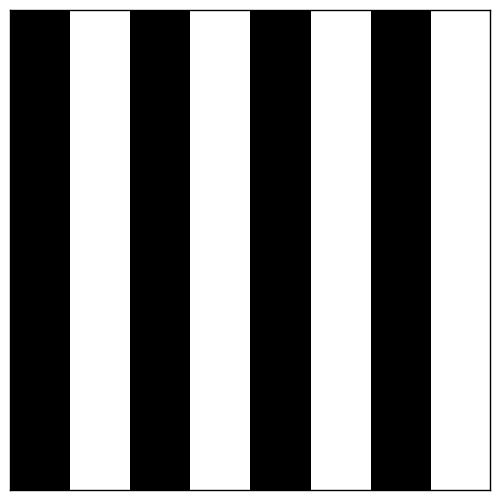} & \ing{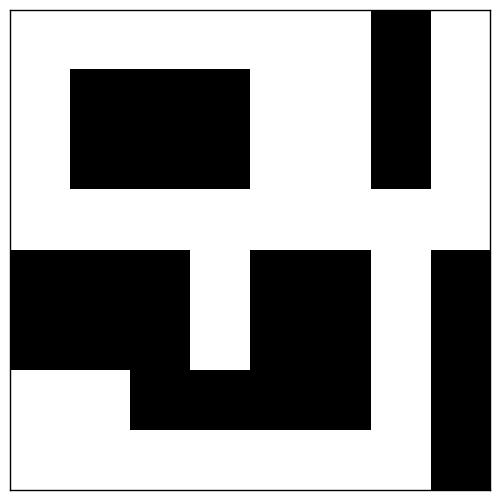} \\  
  };
\end{tikzpicture}
\caption{The algorithm takes in the gray images (row~\ttA) and blurs them to images with $8 \times 8$ resolution (row~\ttB). These are the values sensed by each robot, and are used to calculate the pattern probabilities and choose the most probable local pattern. Row~\ttC shows consensus convergence for the most likely global pattern. For each of the charts in row~\ttC: the $y$ axis shows pattern probability $p$ from 0 to 1, and the $x$ axis shows the number of steps taken from 0 to 35. The red horizontal line marks $p=0.5$. The blurred image from row~\ttB ~is converted to binary (row ~\ttD) to get initial states for the pattern generator, which generates the resultant pattern (row~\ttE).}
\label{fig:8simResults}
\end{figure}

Note that this grid of $8 \times 8$ robots is in many ways analogous to the sensor of a digital camera, albeit a camera with only $8 \times 8$ sensors and thus with very low resolution. If you were to recapture our test images with such a low resolution camera, many different pixels in the test image would contribute to the camera's output, resulting in a very blurry image. We therefore downsample the input image by taking the average of $16 \times 16$ pixel blocks. This blurred image is used as the color sensed by each robot for selecting the most likely pattern. For pattern generation, the initial on/off state is determined by making the blurred image binary (ie. white and black). Figure~\ref{fig:8simResults} shows the entire process for each of the three input images.

Once the Droplets calculate the local pattern based on their sensed color and that of their neighbors, they need to achieve consensus on the global pattern. As has been discussed in Section~\ref{sec:consensus}, convergence of this value is guaranteed.

Next, the pattern generator described above is used (Section~\ref{sec:generator}) with the activator and inhibitor regions seen in Figure~\ref{fig:droplet_morphogen}. The activator field value of $W_1 = 1$ was used, as suggested in paper~\cite{young1984local}. The inhibitor field value, $W_2$, is a parameter which gives rough control over what proportion of the robots are `on' in the final pattern. We found that $W_2 = -0.75$ gave qualitatively good results for all three of the pattern generators.

Finally, we start pattern generation with each robot's initial state to be `on' or `off' status based by the sensed value. If the value is less than $127$ we set it black, otherwise we set it white. The pattern generator runs for ten iterations. 
Robots on the image boundary use a reflection of their neighbors. A robot on the top row, for example, would count its bottom neighbor twice, as the top row is empty.

To further test the simulated algorithm, we added a simple noise model. For measurement error, instead of always assigning the appropriate color to a robot based on its position, we assign a uniformly random color with probability $\rho_{meas}$. For communication error, at each step in the algorithm where information from a neighbor is shared with a robot for a calculation (including the step where a robot's neighbors are calculated in the first place), are robot does not share this information with probability $\rho_{comm}$. 

\begin{figure}[!htb]
\sidecaption[t]
\includegraphics[width=0.6\linewidth]{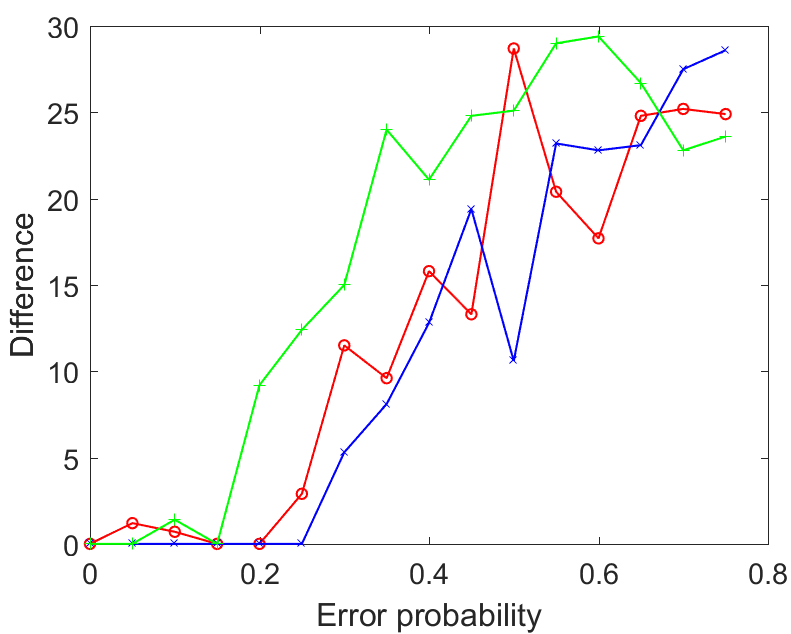}
\caption{The y-axis is the pixel difference from the `correct' pattern and the x-axis is the error probability. The red line shows the effect of measurement errors ($\rho_{meas}$). The blue line shows the effect of communication errors ($\rho_{comm}$). The green line shows the effect of both measurement and communication errors ($\rho_{comm}=\rho_{meas}$). Each data point reflects the mean result over 10 trials of the forest image.}
\label{fig:errorAnalysis}
\end{figure}

For a quantitative measure of the effects of error, we calculated the total absolute difference between the final generated pattern in the presence of error, and the final generated pattern without any error (as visible in the bottom row of Figure~\ref{fig:8simResults}). These results are charted in Figure~\ref{fig:errorAnalysis}. Note that, with the $8x8$ images used, a purely random image should give us a difference of $32$, on average. The algorithm seems quite robust to errors of up to $0.15--0.2$. After these thresholds, the error increases sharply. (Results shown here are for the forest image, with other images yielding similar results.) Qualitatively, we observe that even as errors started to appear, many of the resulting patterns still looked `good', i.e., still had prominent vertical stripes. The main determining factor as the probability of error increased seemed to be in the global pattern detection. If the correct pattern (vertical stripes in this case) is selected, the resulting pattern will fit well even with large errors. Correct pattern selection grows increasingly infrequent, however.

\section{Hardware Implementation}
\label{sec:hardware_imp}

To validate the proposed algorithm and to understand the sorts of errors that real hardware introduces, we implemented the algorithm described above on a swarm of ``Droplets'' ~\cite{farrow2014miniature,klingner2014}. The Droplets are an open-source platform, with source code and manufacturing information available online\footnote{\url{http://github.com/correlllab/cu-droplet}}. Each Droplet is roughly cylindrical with a radius of $2.2\,\mathrm{cm}$ and a height of $2.4\,\mathrm{cm}$.
The Droplets use an Atmel xMega128A3U micro-controller, and receive power via their legs through a floor with alternating strips of $+5V$ and $GND$. Each Droplet has six infrared emitters, receivers, and sensors, which are used for communication and for the range and bearing system~\cite{farrow2014miniature}. The top of each board has sensors to detect the color and brightness of ambient light, and an RGB LED. Each droplet has a 16-bit unique ID.

In our implementation, each Droplet maintains an array of neighbor's IDs. Messages are labeled with phase flags and attached with Droplets' IDs. The Droplets are synchronized using a firefly synchronization algorithm \cite{mirollo1990synchronization,werner2005firefly}. A simple TDMA protocol is used with $37$ slots, each $350$ms long. Each frame is thus $12.95$s long. Each robot is assigned a slot based on its unique id modulo 37. The number of slots (37) was chosen to be large enough that the probability of two adjacent robots sharing a slot is low, but small enough that the algorithm runs quickly.

\begin{figure}[!htb]
\sidecaption[t]
\includegraphics[width=0.4\columnwidth]{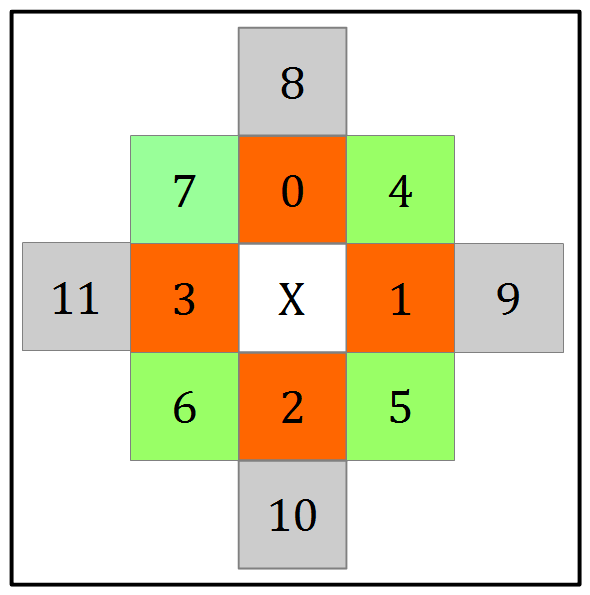}
\caption{Neighbor array. The orange neighbors(0-3) are used for pattern recognition; the green neighbors(4-7) are used in addition to the orange for pattern consensus.All pictured neighbors (orange, green and gray) are used for pattern formation.}
\label{fig:droplet_neighbors}
\end{figure}

In Phase 0 (neighbor identification), we initialize and configure the neighbor ID arrays which store neighboring Droplets' IDs. Range and bearing information is used to calculate positions for each Droplet's immediate neighbors, and neighbors-of-neighbors are learned by listening to the messages sent by Droplets each slot, which contain that Droplet's neighbors. The positions of Droplets and their indices in the array are illustrated in Fig.~\ref{fig:droplet_neighbors}. We allot $20$ frames for Phase 0, since the neighbor information is critical to the three phases.

In Phase 1 (color sensing and recognition), each Droplet communicates the color it senses, and stores the colors its neighbors sense, as learned through communications. Once this is complete, each (non-boundary) Droplet should know the ID and position of 12 neighbors, as well as those neighbor's sensed colors. With this information, the Droplets calculate an pattern probability array $p$ as described in Section~\ref{sec:descriptor}. This phase is allotted $10$ frames.

In Phase 2 (pattern consensus), each Droplet communicates its pattern probability array $p$ and receives pattern probability arrays from its neighbors. At the end of each frame, each Droplet updates its pattern probability array according to the weighted-average consensus algorithm, as described in Section~\ref{sec:consensus}. Each `step' of the consensus algorithm spans one frame. This phase is allotted $35$ frames.

In Phase 3 (pattern formation), each Droplet communicates its intended color for the generated pattern, and receives that information from neighboring Droplets. At the end of each frame, each Droplet updates its color in the generated pattern
from corresponding Droplets. Each Droplet exchanges pattern color message with neighbors. At the end of each frame, each Droplet updates its pattern color according to the pattern generation algorithm described in Section~\ref{sec:generator}.  This phase is allotted $20$ frames.

\section{Hardware Results}
\label{sec:hardware_results}

\begin{figure}[!htb]
\centering
\includegraphics[width=0.3\columnwidth]{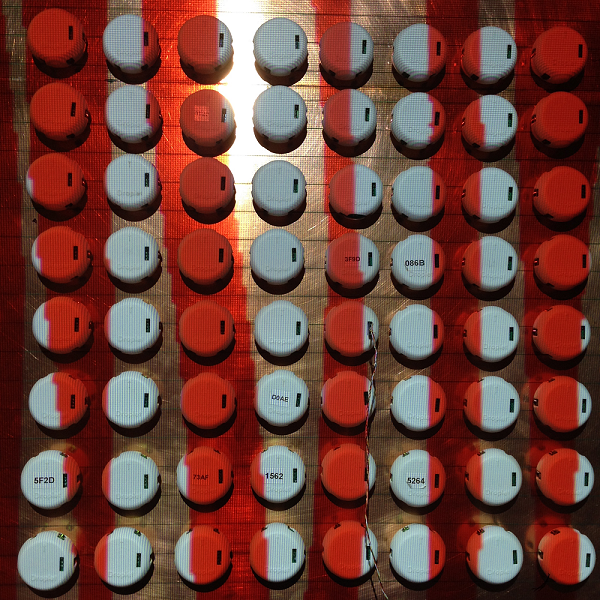}
\includegraphics[width=0.3\columnwidth]{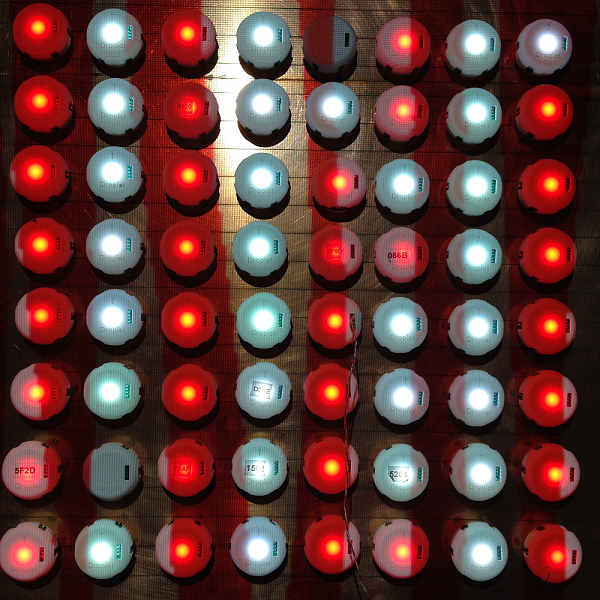}
\includegraphics[width=0.3\columnwidth]{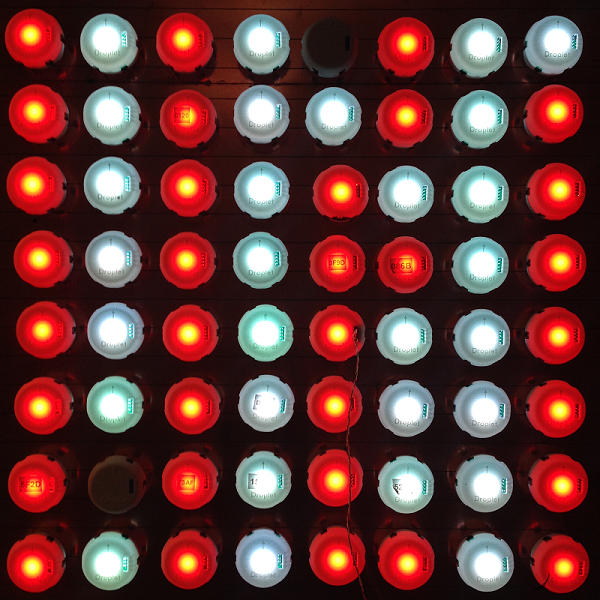}
\caption{Initial condition (left), final pattern with projected pattern (middle) and final pattern (right) for camouflaging the tiger stripe pattern}
\label{fig:t_7_8_11}
\end{figure}

\begin{figure}[!htb]
\sidecaption[t]
\includegraphics[width=0.6\columnwidth]{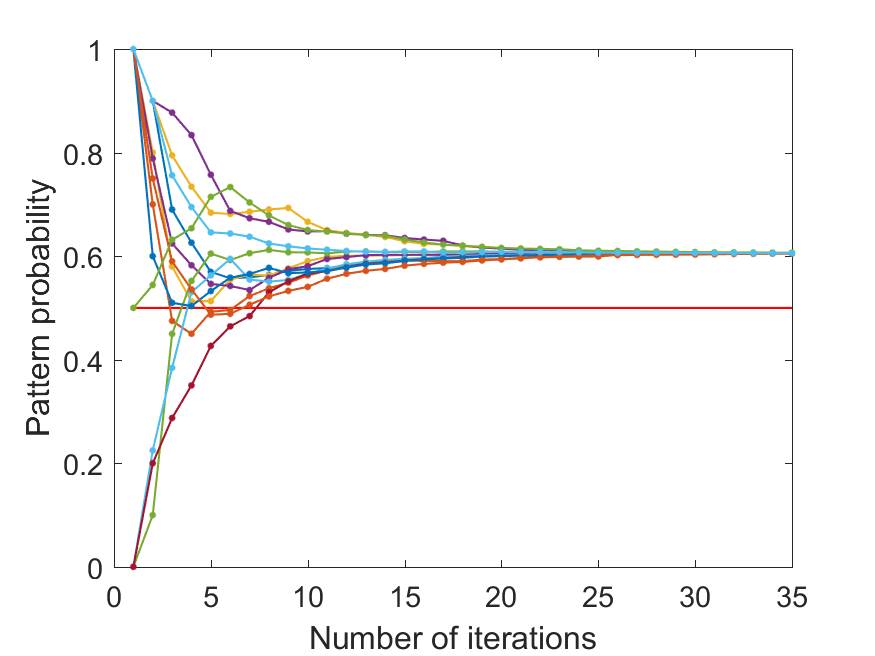}
\caption{Convergence of pattern probabilities of randomly chosen Droplets camouflaging tiger stripe pattern}
\label{fig:t_7_8_11_pp}
\end{figure}

A hardware implementation of the swarm camouflage algorithm is shown in Figure~\ref{fig:t_7_8_11}. For this test, the projected image for the Droplets to sense is a tiger stripe pattern. The results of this test are interesting because a striped pattern is maintained, despite the failure of two units. This, in addition to the more-difficult-to-count failures in communication and color sensing.

Figure~\ref{fig:t_7_8_11_pp} shows the pattern probability convergence for a random sampling of Droplets, when run with a simple horizontal stripe pattern projected on them. The swarm reaches consensus on a horizontal pattern, converging to $p_h=0.61$.

\section{Conclusion}

We present a distributed camouflage system in which a robot swarm can sense the environment color, recognize the local pattern, achieve consensus on the global pattern, and generate a camouflage pattern consistent with the environment the robots are in. In our design, pattern descriptors are proposed for recognizing local patterns. A weighted-average consensus scheme is then utilized, allowing the swarm to converge to a common global pattern. Finally, a pattern formation model is applied to each robot which generates a pattern appropriate for the background. This is accomplished using local communication and simple mathematical operations.

We simulated the proposed algorithm on a couple of patterns from nature: a desert, a forest, and leopard skin. After going through all the phases in the algorithm, and successfully agreeing on a global pattern, the simulation results show that robots with wrong color reading can correct themselves to match the global pattern. This is especially obvious for the horizontal and vertical patterns. We also tried to test the distributed algorithm by applying it on the Droplet swarm robotics platform. The results from the Droplets is promising since the robots can agree on the global pattern and show a proper matching pattern even if individual Droplets stop working. 

As communication on the Droplets is not perfectly reliable, the resultant patterns exhibit some random variations; they do not perfectly match simulation. Even these variations, however, will roughly follow the desired background pattern, seeming to bend or twist around the erroneous robot. In the future, we wish to test the algorithm on a more purpose-built hardware platform which would allow for higher-resolution patterns, and extend the algorithm to include consensus on the dominant colors and patterns consisting of more than two colors.

\begin{acknowledgement}
This research has been supported by NSF grant \#1150223.
\end{acknowledgement}

\bibliographystyle{spmpsci}
\bibliography{reference}

\end{document}